\documentclass[12pt]{elsarticle}

\usepackage{algorithm}% http://ctan.org/pkg/algorithms
\usepackage{algpseudocode}% http://ctan.org/pkg/algorithmicx
\usepackage{amssymb}
\usepackage{amsmath}
\usepackage{float}
\usepackage{hyperref}

\journal{CoRR}

\begin{document}

%    \makeatletter
%    \def\ps@pprintTitle{%
%       \let\@oddhead\@empty
%       \let\@evenhead\@empty
%       \let\@oddfoot\@empty
%       \let\@evenfoot\@oddfoot
%    }
%    \makeatother

\begin{frontmatter}



\title{A predictor-corrector method for the training of deep neural networks}


\author{Yatin Saraiya}

\address{847 Moana Court, Palo Alto, CA 94306, USA}
\ead{yatinsaraiya12@gmail.com}

\begin{abstract}
The training of deep neural nets is expensive.  We present a {\em predictor-corrector} method
for the training of deep neural nets.  It alternates a predictor pass with a corrector pass
using stochastic gradient descent with backpropagation such that there is no loss in validation accuracy.
No special modifications to SGD with backpropagation is required by this methodology.
Our experiments showed a time improvement of 9\% on the CIFAR-10 dataset.
\end{abstract}


\end{frontmatter}


\section{Introduction}
\label{intro}
Image recognition is performed mainly by deep convolutional neural networks \cite{CNNs,backpropagation}, and
the depth of the network is crucial to accurate results \cite{Simonyan}.
However, very deep neural nets degrade near convergence 
 unless some form of shortcutting \cite{bishop} is used.  
We choose the deep residual nets of \cite{residual} as the basis for our work.
These networks consist of stacked blocks with the same number of inputs as 
outputs\footnote{except for 3 changes in input and output sizes.}. The addition of a residual identity mapping
counters the degradation problem.

\section{Methodology}
\label{methodology}

\paragraph{Assumption 1}
Our basic hypothesis is that the weights and biases at the lowest layers of a deep neural network learn more
slowly than those of the upper layers.  Hence, the parameters of the lower layers can be computed every other iteration
with no appreciable loss of validation accuracy. 

\paragraph{Assumption 2}
We also assume that the weights and biases at the lower layers
approximate the identity function. Hence, if network $N_1$ is obtained from $N_2$ by adding some layers at the 
bottom of the stack, then we expect there to be an approximate equivalence in the 
learned parameters of the common (upper) blocks.

\begin{figure*}
\centering
  \includegraphics[width=0.4\textwidth]{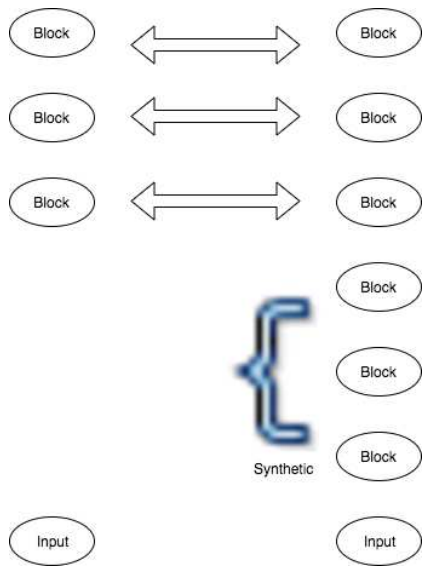}
\caption{Predictor and corrector}
\label{predictor-corrector}       % Give a unique label
\end{figure*}

Our methodology is to maintain and train 2 models, a shallower one (the {\em predictor}) and
a deeper one (the {\em corrector}). At the end of the training, the corrector is the trained model
to be used.  The parameters on all blocks above the input
layer in the predictor are maintained as equal to the parameters of the same number of upper blocks in the corrector.
An alternation with this copy operation is done at the granularity of 1 epoch.
Figure~\ref{predictor-corrector} shows the picture.  The remaining blocks on the corrector are initialized
to be the parameters of the input layer in the case of the input layer, and to the lowest non-input block of
the predictor for the additional blocks of the corrector.  Then SGD and backpropagation maintain these
parameters (i.e they are not copied).

Let $N_1$ be an instance of a resdidual neural net version 1 of \cite{residual}.
Assume it has $L$ blocks, 1 representing the input block and $L$ representing the output block.
Let $B_l$ represent the $l$th block, with parameters $P_l$.  This is the predictor.
Let $N_2$ be the corrector, which is obtained by copying $N_1$ and modifying it as in Algorithm~1  below.
 Note that this is performed only once per training session.
Algorithm~2 describes how to perform the training.  Note that no special processing is required.

\begin{algorithm}
\label{corr}
\caption{Construction of corrector}
\begin{algorithmic}[1]
\Procedure{ConstructCorrector}{$N_1$, $K$}\Comment{$K$ is the number of blocks to add.}
   \State copy $N_1$ to $N_2$
   \For{$i = 1, 2, \ldots, K$}
       \State Add a copy of block $B_2$ in $N_1$ just under $B_2$ in $N_2$
   \EndFor
\EndProcedure
\end{algorithmic}
\end{algorithm}

\begin{algorithm}
\label{pred-corr}
\caption{Training algorithm}
\begin{algorithmic}[1]
\Procedure{Train}{$N_1$, $N_2$, $K$}\Comment{$K$ is the number of blocks added.}
   \For{half the number of epochs}
       \State Perform one training epoch using the predictor $N_1$
       \For{l = 2, 3, \ldots, L}
           \State Copy $P_l$ in the predictor to $P_{l+K}$ in the corrector
       \EndFor
       \State Perform one training epoch using the corrector $N_2$
       \For{l = 2, 3, \ldots, L}
           \State Copy $P_{l+K}$ in the corrector to $P_l$ in the predictor 
       \EndFor
   \EndFor
   \State The corrector $N_2$ is the trained model.
\EndProcedure
\end{algorithmic}
\end{algorithm}

\section{Experiments}

\label{experiments}
\begin{figure*}
\centering
  \includegraphics[width=0.9\textwidth]{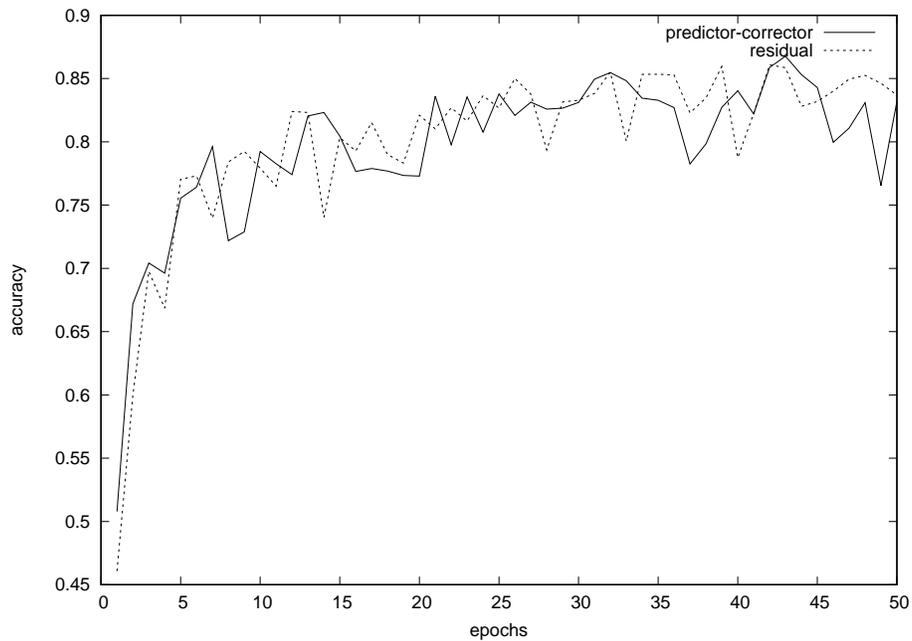}
\caption{Validation accuracy}
\label{residual4}       % Give a unique label
\end{figure*}

\begin{table}
\centering
\label{timestable}       % Give a unique label
\begin{tabular}{|l|c|c|}
\hline\noalign{\smallskip}
&Time savings\\
\noalign{\smallskip}\hline\noalign{\smallskip}
Predictor-corrector residual& 9 \\
\noalign{\smallskip}\hline
\end{tabular}
\caption{Time savings (\%) over 50 epochs}
\end{table}

\begin{table}
\centering
\label{resultstable}       % Give a unique label
\begin{tabular}{|l|c|c|}
\hline\noalign{\smallskip}
&Min top-1 error\\
\noalign{\smallskip}\hline\noalign{\smallskip}
Residual& 14.04 \\
Predictor-corrector residual& 13.24 \\
\noalign{\smallskip}\hline
\end{tabular}
\caption{Top-1 validation error (\%) over 50 epochs}
\end{table}

We used \texttt{cifar10\_resnet.py}, obtained from 
\newline\texttt{https://github.com/fchollet/keras/blob/master/examples/},
to model the experimental framework of Section~4.2 of \cite{residual}.  This software is under the
MIT license.  It is used as the predictor, with 116 layers.
We modified it to create a deeper corrector model by adding 15 layers above the
input layer.
We ran both against the CIFAR-10 dataset \cite{cifar10} for 50 epochs.

The time savings were 9\% (see Table~\ref{timestable}).

\paragraph{Results}
%The time per epoch for the residual net with history was 15\% higher than the time per epoch of the residual net.

Our results are contained in Table~2 and Figure~\ref{residual4}.
Note that the predictor-corrector top-1 validation error is lower than that of the residual net, although marginally so.

\section{Conclusions}
\label{conclusions}
We presented a predictor-corrector methodology for training a deep neural net using alternating
epochs with a shallower and less expensive model.
We gained a time savings of 9\% on the CIFAR-10 dataset with no loss in validation accuracy.





\end{document}